\documentclass{article}

\usepackage{arxiv}

\usepackage[utf8]{inputenc} % allow utf-8 input
\usepackage[T1]{fontenc}    % use 8-bit T1 fonts
\usepackage{url}            % simple URL typesetting
\usepackage{booktabs}       % professional-quality tables
\usepackage{amsfonts}       % blackboard math symbols
\usepackage{nicefrac}       % compact symbols for 1/2, etc.
\usepackage{microtype}      % microtypography
\usepackage{lipsum}		% Can be removed after putting your text content
\usepackage{doi}

\usepackage{amssymb}
\usepackage{gensymb}
\usepackage{multirow}
\usepackage{array,dcolumn}
\usepackage{fixltx2e}
\usepackage{stfloats}
\usepackage{url}
\usepackage{amssymb}
\usepackage{gensymb} 
\usepackage{multirow} 
\usepackage{enumitem}
\usepackage{amsmath}
\usepackage{graphicx}
\usepackage{multirow,bigdelim}
\usepackage{booktabs}
\usepackage{textcomp}
\usepackage{adjustbox}
\usepackage{verbatim}
\usepackage{enumitem}
\usepackage{caption}
\usepackage{subcaption}
\usepackage{mathtools}
\usepackage{tikz}
\usetikzlibrary{graphs, graphs.standard, quotes}
\usepackage{todonotes}
\usepackage{wrapfig}

\usepackage[normalem]{ulem}

\newcommand{\vect}[1]{\boldsymbol{\mathrm{#1}}}
\usepackage{hyperref}   
\usepackage{color}
\title{Bounded nonlinear forecasts of partially observed geophysical systems with physics-constrained deep learning}

\author{%
  Said Ouala$^1$, Steven L. Brunton$^2$, Bertrand Chapron$^3$ \\
  \textbf{Ananda Pascual}$^4$, \textbf{Fabrice Collard}$^5$, \textbf{Lucile Gaultier}$^5$, \textbf{Ronan Fablet}$^1$\\
   $(1)$ IMT Atlantique; Lab-STICC, 29200 Brest, France\\
   $(2)$ University of Washington, USA\\ 
   $(3)$ Ifremer, LOPS, 29200 Brest, France\\
   $(4)$ IMEDEA, UIB-CSIC, 07190 Esporles, Spain\\
   $(5)$ ODL, 29200 Brest, France\\
  \texttt{\{said.ouala, ronan.fablet\}@imt-atlantique.fr}\\
  \texttt{sbrunton@uw.edu},
  \texttt{Bertrand.Chapron@ifremer.fr},\\ \texttt{ananda.pascual@imedea.uib-csic.es} \\
  \texttt{\{dr.fab, lucile.gaultier\}@oceandatalab.com}
}
\renewcommand{\shorttitle}{Bounded nonlinear forecasts of partially observed geophysical systems}

\hypersetup{
pdftitle={Bounded nonlinear forecasts of partially observed geophysical systems with physics-constrained deep learning},
pdfsubject={stat.ML},
pdfauthor={Said Ouala, Steven L. Brunton, Ananda Pascual, Bertrand Chapron,Fabrice Collard, Lucile Gaultier, Ronan Fablet},
pdfkeywords={Partially-observed systems, Embedding, Boundedness, Deep learning, Neural ODE, Forecasting},
}
\begin{document}
\maketitle

\begin{abstract}
The complexity of real-world geophysical systems is often compounded by the fact that the observed measurements depend on hidden variables. These latent variables include unresolved small scales and/or rapidly evolving processes, partially observed couplings, or forcings in coupled systems. 
This is the case in ocean-atmosphere dynamics, for which unknown interior dynamics can affect surface observations. The identification of computationally-relevant representations of such partially-observed and highly nonlinear systems is thus challenging and often limited to short-term forecast applications. Here, we investigate the physics-constrained learning of implicit dynamical embeddings, leveraging neural ordinary differential equation (NODE) representations. 
A key objective is to constrain their boundedness, which promotes the generalization of the learned dynamics to arbitrary initial condition. The proposed architecture is implemented within a deep learning framework, and its relevance is demonstrated with respect to state-of-the-art schemes for different case-studies representative of geophysical dynamics.
\end{abstract}

% keywords can be removed

\section{Introduction}
\label{sec:intro}

Modeling the dynamics of complex systems is one of the key drivers of technological development. 
Dynamical systems provide interpretable models of the evolution of real-world phenomena, enabling understanding and prediction. 
There are a number of practical and theoretical limitations that arise when characterizing and modeling physical phenomenon, 
such as the lack of known governing equations~\cite{brunton_discovering_2016}, the presence of hidden variables that are not measured directly~\cite{bakarji2022discovering}, and difficulties in assessing and enforcing known symmetries and stability~\cite{loiseau2018constrained,kaptanoglu2021promoting,kramer2021stability}.  
% including the existence and utility of governing equations that stem from first physical principles for specific applications such as prediction, control, and data assimilation. 
These limitations motivate the development of new techniques to advance the current state-of-the-art in model discovery.  
The current literature is largely based on optimization techniques~\cite{sun2019optimization} and artificial intelligence (AI) \cite{lecun2015deep} algorithms with the aim of discovering new models, governing equations, conservation laws and symmetries for complex systems \cite{grewal1976identifiability,glover1974parametrizations,wahlberg1988identification,kumpati1990identification,brunton_discovering_2016,patak_PhysRevLett.120.024102,chen2018neural} for a variety of applications, ranging from system identification~\cite{ogunmolu2016nonlinear,brunton_discovering_2016,fablet_bilinear_2017,DuongAssimilationLearning2020,nguyen2020variational,ouala2021learning}, forecasting~\cite{braakmann-folgmann_sea_2017,lim2020time} and reconstruction~\cite{Ouala_2018,gilpin2020deep,fablet2020learning} to control~\cite{cheon2015replacing,kaiser2018sparse}.

One of the new challenges of data-driven methods and learning based approaches concerns the exploitation and development of state-of-the-art techniques to tackle the modeling of complex systems such as the atmosphere, the ocean, and the climate. These fields encountered a sensing revolution in the last years, which sets up the community with databases that enable the use of AI. In this context, significant advances in model reduction techniques \cite{san2018extreme,gupta2021neural}, interpolation and reconstruction of gap-free observations \cite{Ouala_2018,ouala_sea_2018,fablet2020endtoend2joint} and short term forecast \cite{ravuri2021skilful} have been achieved by AI models. These models, when compared to classical physical models, issued for instance from the Navier-Stokes equations, are easier to handle from a computational point of view, they also often offer a more flexible application range with the drawback of been bad learners. AI models often fail generalizing to situations (or initial conditions) that were never seen in the training data. These model also often fail at matching physical model in their ability to simulate the dynamics of the observations. From a pure learning perspective, generalization, and long term simulation performance of data-driven schemes are linked to physical properties of the underlying dynamics (such as conserved quantities) and most of the time, these properties are not revealed by the complexity of the training data.

In this context, a central question concerns the ability of data-driven models to capture the behavior of a system from partial and imperfect observations. Takens theorem~\cite{takens_theorem} states the conditions under which a delay embedding representation of the observations is diffeomorphic to the underlying hidden states. Such a representation, when determined, captures essential information about the hidden states, which makes it possible to learn a model~\cite{bakarji2022discovering}. 
Unfortunately, the derivation of a dynamical system from such a representation remains a challenge, since no explicit relationships between the reconstructed phase space variables and the original hidden variables have been clearly identified. To address this limitation, the Neural embedding of Dynamical Systems (NbedDyn), proposed in~\cite{ouala2020learning}, allows for both reconstructing and forecasting the phase space of the unseen dynamical system. However, this model suffers from generalization and stability issues, especially for initial conditions that are not near the data used to train the model.

It has been shown that embedding prior physical knowledge, such as invariances, symmetries, and conservation laws, into the learning process of dynamical systems may improve the generalizability and stability of the resulting models \cite{loiseau2018constrained,kaptanoglu2020physics,guan2020sparse,wang2020incorporating,wang2020towards,kaptanoglu2021promoting}. 
Within this new field of research, there is an increasing focus on incorporating stability and boundendess constraints into data-driven models. This interest is driven by a combination of works, on different classes of models ranging from sparse regression techniques~\cite{kaptanoglu2021promoting} to deep learning architectures~\cite{manek2020learning,rodriguez2022lyanet}, showing that such constraints can be necessary in order to promote stability and generalization of the models.

Following these developments, we propose a robust framework for the modeling of geophysical dynamical systems from data. This framework relies on the learning of an embedding of the observations governed by a neural ODE (NODE) model \cite{chen2018neural}. This model is constrained to be bounded by restricting the form of the NODE to be linear-quadratic which enables the use of a generalization of the direct Lyapunov method given by the Schlegel boundedness theorem~\cite{schlegel_noack_2015}, which has recently been used to assess the stability of data driven models~\cite{kramer2021stability} and has been incorporated into sparse nonlinear models~\cite{kaptanoglu2021promoting}. On different case-studies, we show that once these constraints are satisfied, the trained representation can reproduce realistic trajectories with respect to the training data using a closed-loop prediction setting. The boundedness constraint also improves the generalization capabilities of the trained model for arbitrary initial conditions, even those lying outside the attractor spanned by the training data. This latter property is extremely important, since the generalization performance is a critical feature for data-driven schemes. Overall, key aspects of the proposed framework are three-fold: 
%\todo[inline]{Lire Please, la partie contributions}
\begin{itemize}
    \item we define a physics-constrained learning-based workflow for the modeling and forecasting of partially observed dynamical systems, with an emphasis on geophysical dynamics;
    \item the proposed framework relies on learning an implicit higher-dimensional embedding. The dynamics of this embedding are constrained to be bounded, which promotes the generalization of models for arbitrary initial conditions, even those far from the training data;
    \item the relevance of the proposed scheme is tested and demonstrated with respect to state-of-the-art methods for short-term forecasting and long-term simulation on several case studies, namely the Lorenz 63 and Lorenz 96 systems and the shallow water equation dynamics. Specifically, our model is compared to the sparse regression (SR) technique \cite{brunton_discovering_2016}, to a recurrent neural network (RNN), to the Latent-ODE model as presented in \cite{chen2018neural} and to an extended dynamic mode decomposition (EDMD) technique based on delay embedding observables \cite{kamb2020timeDobsKoopman}. Through all experiments, the proposed framework is the only model able to achieve simultaneously both an accurate short-term forecast and long-term simulation of the dynamics.% highlighted for instance through the analysis of the spatio-temporal spectrums 
\end{itemize}
Regarding the data-driven identification of real-world geophysical dynamics, we believe that this proposed framework provides a promising approach for learning consistent models in terms of short-term and long-term forecasts through the implementation of physical constraints from prior knowledge of the conservation laws governing the dynamics.

The paper is organized as follows. We start in section \ref{sec:related_works} by introducing state-of-the-art literature on data-driven dynamical systems and physics based learning. We then briefly discuss, in section \ref{sec:SOTA}, stability criteria, both from a dynamical systems theory and an identification perspective. We also introduce a generalization of the direct Lyapunov theorem as proposed in~\cite{schlegel_noack_2015}. Section \ref{sec:ProbFormulation_and_method} presents the proposed framework, followed by the experiments and results in Section \ref{sec:Num_exp}. We close this paper with conclusions and perspectives for future works in Section \ref{sec:Conclusion}.

\section{Related work}
\label{sec:related_works}
\textbf{Data-driven identification of dynamical representations.} Stated above, recent advances in learning-based approaches motivate numerous investigations to apply these tools for the identification of dynamical representations from data. A full review of state-of-the-art techniques is out of the scope of this paper, and we concentrate on providing a high-level overview, organized by the nature of the training data. Specifically, when provided with observations that form an embedding of a deterministic time-evolving system, several works including symbolic or dictionary-based approaches \cite{Schmidt81,paduart_identification_2010,PhysRevLett.106.154101,brunton_discovering_2016,yuan2019data}, neural-network and deep learning approaches \cite{wiewel2018latentspace,raissi2019physics, chen2018neural,Res_INN} and model-free representations \cite{shen_all_seq_seq_pred,patak_PhysRevLett.120.024102} demonstrated the ability of data-driven techniques to reverse engineer the governing equations from a sequence of observations.

When provided with noisy and sparse observations, studies have investigated either deep learning generative architectures, such as variational autoencoders \cite{he_deep_2015,fraccaro_sequential_2016,krishnan_structured_2016}, or classical Bayesian filtering schemes such as the ensemble Kalman filter \cite{evensen2009data}, in deriving reconstructions of the irregular observations that can be useful from an identification perspective \cite{bocquet2019data,brajard2020combining,nguyen2020variational}.

When provided with partial observations, {\em i.e.} with no one-to-one mapping between the observation space and the underlying dynamical system, most state-of-the-art techniques do not apply since the variability of the observations is influenced by some missing components, making the temporal evolution of the observations stochastic in the observation space. From this point of view, the deep learning literature in time series forecasting mainly relies on some sort of recurrent neural network (RNN) architectures \cite{kumar2004river,kumar2018energy,tsai2018air} which can be motivated from a dynamical system perspective as a temporal propagator of a delay embedding of the observations. 
Recent work has connected SINDy with time-delay autoencoders to reconstruct hidden variables, with promising results~\cite{bakarji2022discovering}. 

Methods to learn linear propagators of non-linear, and potentially high-dimensional dynamical systems have also emerged in the context of the modern Koopman theory \cite{KoopmanGeneral}. Finding a suitably chosen transformation/embedding of the observations that can be approximately propagated linearly in time has motivated several works \cite{williams2015data,brunton2016koopman,brunton2017chaos,takeishi2017learning,lusch2018deep,yeung2019learning,kamb2020timeDobsKoopman,lange2020fourierForecast,azencot2020forecasting,rice2020analyzing} on a variety of applications, including control \cite{mauroy2020koopman} and prediction \cite{lange2020fourierForecast}.

\textbf{Physics-informed data-driven dynamical representations.}  In the context of dynamical model identification, several works have shown that including physical knowledge has a positive impact on the data-driven models~\cite{majda2012physics,loiseau2018constrained,raissi2019physics,mardt2020deep,wang2020incorporating,ma2020combined,guan2020sparse,wang2020towards,jin2021nsfnets,cai2021physics,kashinath2021physics,mahmoudabadbozchelou2021data,kharazmi2021hp,yang2021b}. For instance, \cite{brunton_discovering_2016} proposed a sparse regression framework for the derivation of interpretable dynamical representations from data and \cite{loiseau2018constrained} extended this framework to satisfy physical constraints, such as known energy preserving symmetries in the dynamics. 
In neural-networks and deep learning, significant effort has been focused on understanding residual networks \cite{he2016deep,he2016identity} as numerical integration schemes of differential equations \cite{weinan2017proposal,fablet_bilinear_2017,Res_INN,ruthotto2020deep,rousseau2020residual}. In this line of study, neural ordinary differential equations \cite{chen2018neural,dupont2019augmented,yan2019robustness,Acc_rate_grad_NODE1,Acc_rate_grad_NODE2,tuor2020constrained} have shown great success in merging neural networks and ODE identification techniques. 

There is also a growing interest in developing hybrid representations that merge both a physical model and a machine learning component. Typical examples can be found in the context of closure modeling \cite{ling2016reynolds,BECK2019108910,zanna2020data,frezat2020physical,charalampopoulos2021machine,gupta2021neural} in which, broadly speaking, a machine learning model is optimized on data to correct a physical model.

\textbf{Stability and boundedness of data-driven dynamical systems.}   Exploiting stability constraints when learning dynamical systems from data usually improves the long-term stability and generalization performance of the trained models \cite{neumann2013neural,benosman2016learning,mamakoukas2020memory,wang2020deep,kaptanoglu2021promoting}. Recent works further investigated Lyapunov functions of dynamical systems for control applications \cite{taylor2019episodic} or to force the stability around a global equilibrium point \cite{manek2020learning}. In this context, \cite{rodriguez2022lyanet} proposed a new framework that guarantees exponential stability of an NODE model. Promoting Lyapunov stability in autoencoders was also investigated in \cite{erichson2019physics} with applications to reduced-order models of two-dimensional turbulent fluid flows. Beyond the direct Lyapunov method, the Schlegel boundedness theorem \cite{schlegel_noack_2015} was also considered in \cite{kaptanoglu2021promoting} to promote the global stability of linear quadratic data-driven dynamical systems.

\section{From Stability to Boundedness, in the context of identification}
\label{sec:SOTA}
This section introduces some common stability criteria for the analysis of non-linear models. We briefly discuss the implementation of these criteria and focus on boundedness constraints as we further explore their relevance to regularize the learning of chaotic dynamics.

\subsection{Stability of limit-sets}
Let us assume a continuous $s$-dimensional dynamical system $\vect{z}_t$ governed by an autonomous ODE $\Dot{\vect{z}}_{t} = f(\vect{z}_{t})$ with $\Phi_t$ the corresponding flow $\Phi_t(\vect{z}_{t_0}) = \vect{z}_{t_0}+\int_{t_0}^{t}f(\vect{z}_{w})dw$. The trajectories of this dynamical system are assumed to be asymptotic to a limit-set $\mathcal{S}$ of dimension ${p}$ contained in $\mathbb{R}^s$.

Stability theory addresses the characterization of the asymptotic behavior of a set of solutions of a differential equation with respect to a given limit-set $\mathcal{S}$. Formally, and assuming for the sake of simplicity that $\mathcal{S}$ is an equilibrium point $\vect{z}_{eq}$, we may distinguish the following stability definitions:

\begin{itemize}
\item The equilibrium point $\vect{z}_{eq}$ is globally (respectively locally) stable if $\forall \vect{z}_{0} \in \mathbb{R}^s$ (respectively $\vect{z}_{0} \in \mathcal{U} \subset \mathbb{R}^s$), $\Phi_t(\vect{z}_{0}) \longrightarrow \vect{z}_{eq}+\epsilon$ with  $ |\epsilon|>0$ and finite;

\item The equilibrium point $\vect{z}_{eq}$ is globally (respectively locally) asymptotically stable if $\forall \vect{z}_{0} \in \mathbb{R}^s$ (respectively $\vect{z}_{0} \in \mathcal{U} \subset \mathbb{R}^s$), $\Phi_t(\vect{z}_{0}) \longrightarrow \vect{z}_{eq}$ as $t \longrightarrow \infty$;

\item The equilibrium point $\vect{z}_{eq}$ is globally (respectively locally) exponentially stable if $\forall \vect{z}_{0} \in \mathbb{R}^s$ (respectively $\vect{z}_{0} \in \mathcal{U} \subset \mathbb{R}^s$), $|\Phi_t(\vect{z}_{0}) - \vect{z}_{eq}| \leq C|\vect{z}_{0}- \vect{z}_{eq}|e^{-\alpha t}$ with $\alpha$ a convergence rate, $C$ a positive constant and $|\cdot |$ a given norm in $\mathbb{R}^s$.
\end{itemize}
Broadly speaking, we may state that a limit-set is i) stable if two nearby trajectories stay close by the action of the vector field, ii) asymptotically stable if a sufficiently close trajectory is attracted to the limit-set, and finally iii) exponentially stable if a trajectory converges to the limit-set with an exponential decay rate. 

Depending on the limit-set of a given ODE, there are several approaches to analyze its stability. A great introduction to classical stability criteria is given in \cite{Parker1989}, starting from the classical eigenvalues of a linear (or linearized) system around an equilibrium point and finishing with the Lyapunov exponents of chaotic attractors. From an identification perspective, and given some observation data, stability criteria can be used as constraints, either to reproduce an observed asymptotic behavior or to avoid blowups, as long as the attractor revealed by the observations is not strange {\em i.e.} chaotic. Chaotic solutions of differential equation are only revealed through criteria that exploit long-term simulations of the system such as Lyapunov exponents. Therefore, they cannot be characterized based on the dynamical equation (such as the eigenvalues of an equilibrium point). 

Overall, one cannot explicitly constrain a system to be chaotic, simply by looking or manipulating its differential equation. This point appears critical, since numerous geophysical systems can be chaotic. Alternatively, we propose to relax the problem by moving from constraining chaos to constraining the boundedness of a system. Such a boundedness constraint applies to every single limit-set mentioned above and would avoid blowups of data-driven models. As highlighted by \cite{Parker1989}, this constraint is also a natural feature of real-world geophysical systems as, from an experimental perspective, blowups cannot be observed in nature. For this purpose, we start by introducing the direct Lyapunov stability criterion, since it provides global stability properties on non-linear dynamics. A generalization of this approach to the boundedness of linear quadratic ODEs as proposed in \cite{schlegel_noack_2015} is then presented. 

\subsection{Lyapunov stability of dynamical systems}
The direct Lyapunov stability method \cite{lyapunov1992general} was introduced to study the stability of any dynamical system that admits an equilibrium point at the origin. It uses a scalar function of the state $V: \mathbb{R}^s \longrightarrow \mathbb{R}$ as follows:
\begin{equation}
\begin{aligned}
&V(\vect{z}) = 0  \mbox{ if and only if } \vect{z} = 0\\
&V(\vect{z}) > 0\mbox{ if and only if } \vect{z} \neq 0\\
&\Dot{V}(\vect{z})\mbox{ }\leq 0\mbox{ }\forall\vect{z} \neq 0
\label{eq:C5_Lyap_function}
\end{aligned}
\end{equation}
If $V(\vect{z})$ satisfies the above conditions  $\forall \vect{z} \in \mathbb{R}^s$ (respectively $\forall \vect{z} \in \mathcal{U} \subset \mathbb{R}^s$) the system is globally (respectively locally) stable. Furthermore, if $\Dot{V}(\vect{z}) < 0 \forall \vect{z} \neq 0$ the asymptotic stability of the system is also guaranteed.

We can apply this criterion to any non-linear model without resorting to any linearization, which makes it particularly appealing. Yet, finding an appropriate function is far from being straightforward, and several works proposed candidate Lyapunov functions for various types of problems \cite{malisoff2009constructions}.

\subsection{Generalization to boundedness of LQM}

The direct Lyapunov stability method is restricted to dynamical systems with an equilibrium point at the origin. This property is restrictive since it does not apply to other dynamical regimes such as periodic and chaotic solutions. Furthermore, the choice of the Lyapunov function being non-systematic, the exploitation of a direct Lyapunov constraint for the data-driven identification of dynamical systems  is restricted to a small class of parametric models. Fortunately, Schlegel and Noack~\cite{schlegel_noack_2015} proposed a generalization of the direct Lyapunov method on a class of parametric differential equations. This work also introduces  the associated Lyapunov function, along with a condition for the existence of a monotonically attracting trapping region in the phase space. 
These trapping regions were later incorporated into nonlinear model identification to improve stability~\cite{kaptanoglu2021promoting}.

A trapping region is a domain in the phase space that locks trajectories of a dynamical system, {\em i.e.}, once a trajectory enters a trapping region, it will stay in this domain as the system evolves \cite{meiss2007differential}. When this region is globally monotonically attractive, all trajectories in the phase space will converge to the trapping region. We may point out that a trapping region can contain a single or multiple limit-sets. The class of models for which the proposed criterion in \cite{schlegel_noack_2015} is valid are linear quadratic models (LQMs). They can be encountered, for instance, by a spectral discretization of the Navier-Stokes equation, which make them appealing in the context of reduced order models (ROMs). Formally, as proposed in \cite{schlegel_noack_2015}, let us rewrite the dynamical system governing $\vect{z}$ as a linear quadratic model
\begin{equation}
  \Dot{\vect{z}}_{t} = \vect{c} + \vect{L}\vect{z}_t + [\vect{z}^T_t \vect{Q}^{(1)}\vect{z}_t, ..., \vect{z}^T_t\vect{Q}^{(s)}\vect{z}_t]^T
  \label{eq:C5_LQM}
\end{equation}
where $\vect{c} \in \mathbb{R}^s$, $ \vect{L} \in \mathbb{R}^{s \times s}$ and $\vect{Q}^{(i)} = [q_{i,j,k}]^s_{j,k=1},i = 1, ..., s$. The $s$ symmetric quadratic matrices $\vect{Q}^{(i)}$, $i = 1, ..., s$ are supposed to be energy preserving, {\em i.e.} \begin{equation}
  q^{i}_{j,k}+q^{j}_{i,k}+q^{k}_{i,j} = 0, i,j,k = 1, ..., s
  \label{eq:C5_Energy_pres_cond}
\end{equation} 

Let us also consider a shifted variable $\Bar{\vect{z}} = \vect{z}-\vect{m}$ with $\vect{m} \in \mathbb{R}^s$ an arbitrary finite state. The dynamical equation of the shifted state can be written as 
\begin{equation}
  \Dot{\Bar{\vect{z}}}_{t} = \vect{d} + \vect{A}\Bar{\vect{z}}_t + [\Bar{\vect{z}}^T_t\vect{Q}^{(1)}\Bar{\vect{z}}_t, ..., \Bar{\vect{z}}^T_t\vect{Q}^{(s)}\Bar{\vect{z}}_t]^T
  \label{eq:C5_shifted_LQM}
\end{equation}
with
\begin{equation}
  \vect{d} = \vect{c} + \vect{L}\vect{m} +  [{\vect{m}}^T\vect{Q}^{(1)}{\vect{m}}, ..., {\vect{m}}^T\vect{Q}^{(s)}{\vect{m}}]^T%(c_i + \sum^s_{j = 1} l_{ij}m_j+ \sum^s_{j,k = 1} q_{ijk}m_jm_k)^s_{i = 0}
  \label{eq:C5_shifted_bias_true_ODE}
\end{equation}
and 
\begin{equation}
  \vect{A} = \bigg( a_{ij} \bigg)= \bigg( l_{ij} + \sum^s_{k = 1}(q_{i,j,k}+q_{i,k,j})m_k \bigg)
  \label{eq:C5_shifted_linear_true_ODE} 
\end{equation}

The evolution of the fluctuation energy $K = \frac{1}{2} \sum_{i=1}^{s} \Bar{\vect{z}}_i^2$ of the shifted system is considered as a Lyapunov function \cite{schlegel_noack_2015}. The time derivative of this quantity can be written as: 
\begin{equation}
\dot{K} = [\nabla_{\Bar{\vect{z}}} \vect{K}]^T \dot{\Bar{\vect{z}}} = \Bar{\vect{z}}^T \vect{A}_s\Bar{\vect{z}} + \vect{d}^T\Bar{\vect{z}}
\label{eq:C5_shifted_Lyap_func}
\end{equation}
where $\vect{A}_s = \frac{1}{2}(\vect{A} + \vect{A}^T)$. The contribution of the quadratic terms to the fluctuation energy $K$ is zero due to the energy-preserving condition \eqref{eq:C5_Energy_pres_cond}. A sufficient condition for the existence of a monotonically attracting trapping region is the existence of a finite $\vect{m}$ such that $\vect{A}_s$ has only negative eigenvalues. In the next section, we enforce this condition in the proposed data-driven scheme for partially-observed systems in order to train long-term bounded dynamical models.

\section{Bounded Neural Embedding}
\label{sec:ProbFormulation_and_method}

This section first briefly outlines the neural embedding of dynamical systems ---NbedDyn---\cite{ouala2020learning}, and  introduces the bounded neural embedding framework. 
\subsection{Neural embedding of dynamical systems}
Let us consider a dynamical system governed by an autonomous ODE:
\begin{equation}
\Dot{\vect{z}}_{t} = f(\vect{z}_{t}) 
\end{equation}
For most applications, the true state $\vect{z}_t \in \mathbb{R}^s$ of the system is unknown and limited to  a series of observations $\{ \vect{x}_t \}$:
\begin{equation}
\vect{x}_t =\mathcal{H}(\vect{z}_t)
\end{equation}
where $\mathcal{H}: \mathbb{R}^s \rightarrow \mathbb{R}^n$ is an observation operator that does not satisfy the conditions \cite{Sauer1991} under which the predictable deterministic dynamics expressed in the space of $\vect{z}$ are still deterministic in the observation space.

The NbedDyn technique tackles this problem by searching an augmented space, where the states are governed by diffeomorphic flows and can be mapped to the observations $\vect{x}_t$. For any given operator $\mathcal{H}$ of a deterministic dynamical system, Takens theorem \cite{takens_theorem} guarantees that such an augmented space exists. However, instead of using a delay embedding, NbedDyn defines a $d_E$-dimensional augmented space with states $\vect{u}_t \in \mathbb{R}^{d_E}$ as follows:
\begin{equation}
\vect{u}_t^T = [\mathcal{M}(\vect{x}_{t}^T), \vect{y}_{t}^T]
\label{eq:C5_aug_vect}
\end{equation}
where $\mathcal{M}$ is an order reduction operator such as $\mathcal{M}(\vect{x}_{t}) \in \mathbb{R}^r$ with $r \leq n$ and $\vect{y}_{t} \in \mathbb{R}^{d_E-r}$ are stated as the latent states. They account for the information of the unobserved components of the true state $\vect{z}_t$.

The corresponding dynamics and observation operator are defined as:
\begin{align}
\dot{\vect{u}}_t &=f_{\theta}(\vect{u_t}) \label{eq:C5_aug_ode} \\
    \vect{x}_t   &=\mathcal{M}^{-1}(\vect{G}\vect{u_t})
\label{eq:C5_aug_obs}
\end{align}
where the dynamical operator $f_{\theta}$ belongs to a given family of neural network operators parameterized by a parameter vector ${\theta}$. $\vect{G}$ is a projection matrix that satisfies $\mathcal{M}(\vect{x}_t) = \vect{G}\vect{u_t}$. Using an integration scheme (in all the experiments, the flow of the proposed framework is built on a classical Runge-Kutta 4 integration scheme), we can associate $f_{\theta}$ with a diffeomorphic mapping, to derive the flow $\Phi_{\theta}$ of the NODE model \eqref{eq:C5_aug_ode} as follows:
\begin{equation}
\Phi_{\theta,t}(\vect{u}_{t-1}) = \vect{u}_{t-1} + \int_{t-1}^{t}f_\theta (\vect{u}_{t-1})
\label{eq:C5_aug_int}
\end{equation}
From Eqs. (\ref{eq:C5_aug_ode}), (\ref{eq:C5_aug_obs}) and (\ref{eq:C5_aug_int}), we define a state space model:
\begin{equation}
\
\left \{
\begin{array}{ccl}
    \vect{u}_{t} & = &\Phi_{\theta,t}(\vect{u}_{t-1}) \\
    \vect{x}_t & = & \mathcal{M}^{-1}(\vect{G}\vect{u_t})
\end{array}
\right.
\normalsize
\end{equation}
Given an observation time series of size $N+1$ $\{\vect{x}_0,\ldots,\vect{x}_{t_N}\}$, the Neural Embedding scheme aims to minimize the forecasting error of the observations with respect to the model parameters and the latent states as follows:
\begin{gather}
 \hat{\theta}, \vect{y}_{t_0:t_{N-1}} = \displaystyle \arg \displaystyle \min_{\theta,\{\vect{y}_{t}\}_{t}} \sum_{i=1}^{N} \| \vect{x}_{t_i} - \mathcal{M}^{-1}(\vect{G} \Phi_{\theta,t_i} \left(\vect{u}_{t_{i-1}}) \right ) \|^2
+ \lambda_1 \| \vect{u}_{t_i} -  \Phi_{\theta,t_i}(\vect{u}_{t_{i-1}})\|^2
\label{eq:C5_opti_crit 2}
%\raisetag{25pt}
\end{gather}
with $\lambda_1$ a weighting parameter and $\| \cdot \|$ the $L^2$ norm. 

\subsection{Constrained dynamical embedding}

\begin{figure*}
\centering
\includegraphics[width=1.0\textwidth,height=9.0cm]{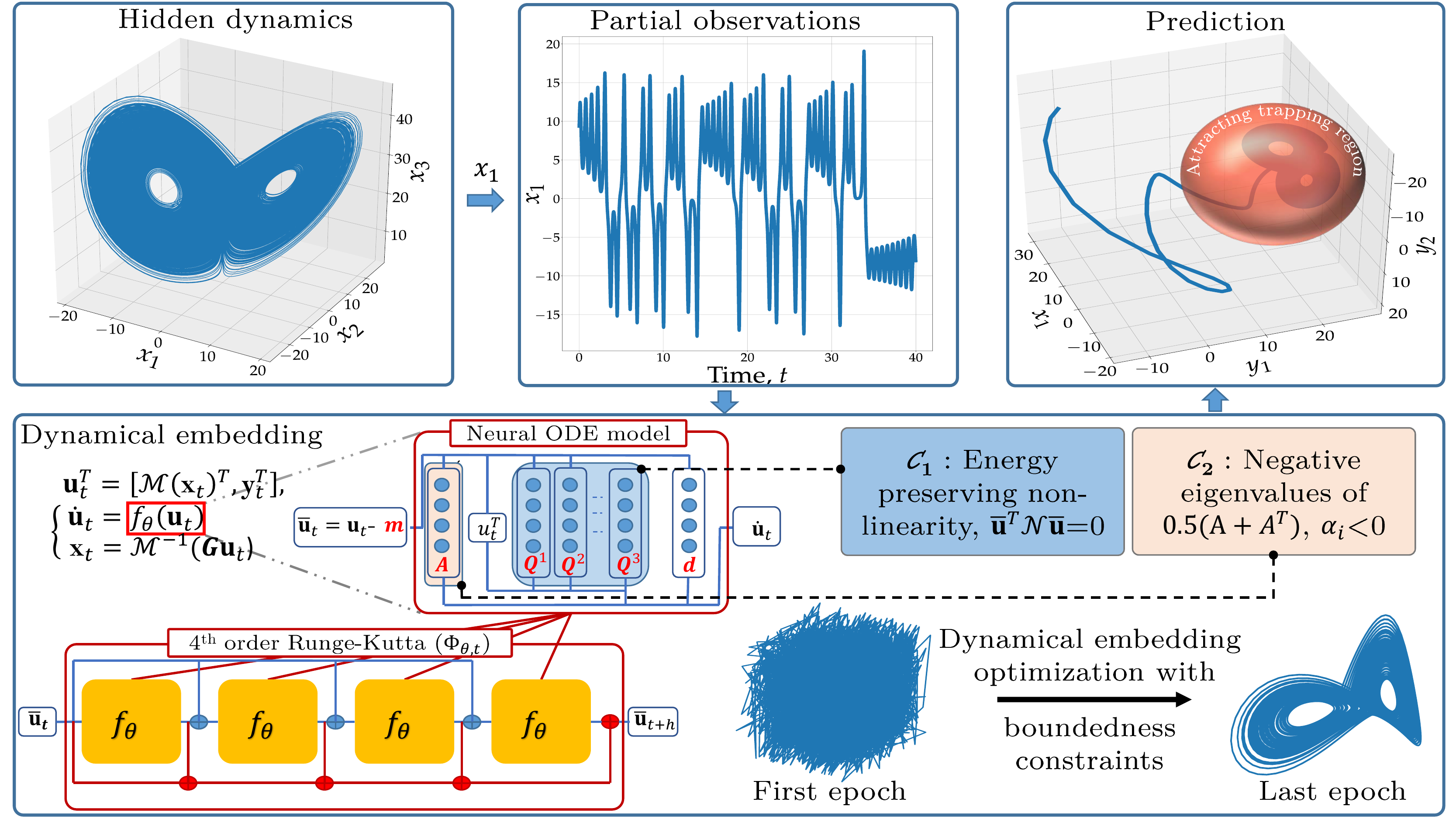}
\caption{{{ \bf   \em Sketch of the proposed architecture.} Given a sequence of observations issued from an unknown higher dimensional dynamical system, the observations are projected into a higher dimensional space parameterized by a linear quadratic neural ODE model. The parameters of the models, highlighted in red, as well as the latent states are optimized to minimize the forecasting of the observations while satisfying the boundedness constraints as proposed in \cite{schlegel_noack_2015}.}}
\label{fig:sketchProp}
\end{figure*}

When considering the parametrization of the model $f_{\theta}$, a linear-quadratic formulation is suitable for the identification of reduced order models of incompressible flows \cite{holmes2012turbulence}. Furthermore, this architecture makes the application of the attracting trapping region condition, introduced in the previous section, tractable using the Schlegel boundedness theorem \cite{schlegel_noack_2015}. Regarding the approximation properties of this class of models, the combination of Volterra theory, on the solutions of non-linear differential equations \cite{langer1932vito}, with lifting transformation techniques \cite{VolterraQLROM2011,qian2020lift} shows that linear quadratic models can faithfully represent (in a higher dimension) the dynamics of non-linear differential equations. In this context, the operator $f_\theta$ in \eqref{eq:C5_aug_ode} is restricted to be linear quadratic, and can be written, similarly to \eqref{eq:C5_LQM} as follows: 

\begin{equation}
  \Dot{\vect{u}}_{t} = \vect{c} + \vect{L}\vect{u}_t + [\vect{u}^T_t\vect{Q}^{(1)}\vect{u}_t, ..., \vect{u}^T_t\vect{Q}^{({d_E})}\vect{u}_t]^T
  \label{eq:C5_LQM_approx}
\end{equation}
where $\vect{c} \in \mathbb{R}^{d_E}$, $ \vect{L} \in \mathbb{R}^{{d_E} \times {d_E}}$ and $\vect{Q}^{(i)} = [q_{i,j,k}]^{d_E}_{j,k=1},i = 1, ..., {d_E}$. The above approximate model is shifted according to $\bar{\vect{u}} = \vect{u}-\vect{m}$ with $\vect{m} \in \mathbb{R}^{d_E}$. The approximate dynamical equation of the shifted state can be written as 
\begin{equation}
  \Dot{\bar{\vect{u}}}_{t} = \vect{d} + \vect{A}\bar{\vect{u}}_t + [\bar{\vect{u}}^T_t\vect{Q}^{(1)}\vect{u}_t, ..., \bar{\vect{u}}^T_t\vect{Q}^{({d_E})}\bar{\vect{u}}_t]^T
  \label{eq:C5_shifted_LQM_hat}
\end{equation} with $\vect{d}$ and $\vect{A}$ computed according to \eqref{eq:C5_shifted_bias_true_ODE} and \eqref{eq:C5_shifted_linear_true_ODE} respectively.

The training setting comes to jointly learning the model parameters $\theta = \{\vect{c}, \vect{L}, \vect{Q^{1}}, \vect{Q^{2}}, \cdots, \vect{Q^{d_E}}, \vect{m}\}$ and the latent states $\vect{y}$ according to the following constrained optimization problem
\begin{gather}
\begin{aligned}
 \hat{\theta}, \vect{y}_{t_0:t_{N-1}} = \displaystyle \arg \min_{\theta,\{\vect{y}_{t}\}_{t}}
& \displaystyle \sum_{i=1}^{N} \| \vect{x}_{t_i} - \mathcal{M}^{-1}(\vect{G} (\Phi_{\theta,t_i} \left(\vect{u}_{t_{i-1}})+\vect{m}) \right ) \|^2 \\
    &+ \lambda_1 \| \vect{u}_{t_i} -  \Phi_{\theta,t_i}(\vect{u}_{t_{i-1}})\|^2 \\~\\
    &+ \lambda_2 \mathcal{C}_1  \\~\\
    &+ \lambda_3 \mathcal{C}_2 
\end{aligned}
\label{eq:C5_opti_crit_const}
\raisetag{50pt}
\end{gather}
with $\mathcal{C}_1 = \sum_{i,j,k=1}^s \|q_{i,j,k}+q_{i,k,j}+b_{j,i,k}+b_{j,k,i}+b_{k,i,j}+b_{k,j,i}\|^2$ and $\mathcal{C}_2 = \sum_{i=1}^s  \mathrm{Max}(\alpha_i,0)/\mathrm{Max}(\alpha_i+1,0)$ where $\alpha_i, i =1, ..., {d_E}$ the eigenvalues of the matrix $\vect{A}_s = \frac{1}{2}(\vect{A} + \vect{A}^T)$. The variables $\lambda_{1,2,3}$ are constant weighting parameters. This loss function corresponds to the initial NbedDyn loss given by \eqref{eq:C5_opti_crit 2} with two additional constraints $\mathcal{C}_1$ and $\mathcal{C}_2$. The first constraint $\mathcal{C}_1$ steams from the energy-preserving condition given in \eqref{eq:C5_Energy_pres_cond}. It forces the contribution of the quadratic terms of $f_\theta$ to the fluctuation energy to sum up to zero. The second constraint, $\mathcal{C}_2$, ensures that the eigenvalues of $\vect{A}_s$ are negative. Satisfying these constraints guarantees that the model $f_{\theta}$ is bounded through the existence of a monotonically attracting  trapping region that includes the limit-set revealed by the minimization of the forecasting loss. Overall, Fig. \ref{fig:sketchProp} highlights the main mechanisms of the proposed framework.

\section{Numerical experiments}
\label{sec:Num_exp}
Numerical experiments are performed to evaluate and demonstrate the relevance of the proposed framework for the forecasting and simulation of partially-observed systems. We consider three case-studies: the Lorenz-63, Lorenz-96, and Shallow Water Equation (SWE) dynamics. The code associated to this work can be found at \href{url}{https://github.com/CIA-Oceanix/Bounded-NbedDyn}.

\subsection{Lorenz 63}
The Lorenz-63 dynamical system is a 3-dimensional model governed by the following ODE: 
\begin{equation}
\label{eq:lorenz-63}
\left \{\begin{array}{ccl}
\frac{dz_{t,1}}{dt} &=&\sigma \left (z_{t,2}-z_{t,2} \right ) \\
\frac{dz_{t,2}}{dt}&=&\rho z_{t,1}-z_{t,2}-z_{t,1}z_{t,3} \\
\frac{dz_{t,3}}{dt} &=&z_{t,1}z_{t,2}-\beta z_{t,3}
\end{array}\right.
\end{equation}
Under parametrization $\sigma =10$, $\rho=28$ and  $\beta=8/3$, this system exhibits chaotic dynamics with a strange attractor \cite{lorenz_deterministic_1963}.

We simulate a Lorenz-63 state sequence of $5000$ time steps, sampled regularly at $dt = 0.01$. This simulation is computed using the LOSDA ODE solver \cite{odepack}. The training sequence consists of the initial $4000$ measurements of the first Lorenz-63 variable, {\em i.e.}, $\mathcal{H} = [1,0,0]$ and $\vect{x}_t = \mathcal{H}(z_t) = \vect{z}_{t,1}$. The remaining time steps are used as a test set.  In this experiment, the projection operator in \eqref{eq:C5_aug_vect} is the identity matrix, {\em i.e.} $\mathcal{M(\cdot)} = I_{1}$. 

\textbf{Parameterization of the data-driven models}: For benchmarking purposes, we perform a quantitative comparison with other leading approaches using delay embedding representations \cite{takens_theorem}. The parameters of the delay embedding representation, namely the lag $\tau$ and the dimension $d_E$ of the augmented space, were chosen using standard techniques. Specifically, the lag parameter was computed using both the mutual information and correlation  techniques \cite{takens_params_1}, respectively denoted as $\tau_{MI}$ and  $\tau_{Corr}$. Regarding the dimension of the embedding representation, we used the Whitney's embedding condition $d_{E} = 2\mathrm{p}+1$ with $\mathrm{p}$ the dimension of the hidden limit-set. The delay embedding dimension was also computed using the false nearest neighbors (FNN) method \cite{Takens_params_2}. We also tested arbitrary parameters for the delay embedding dimension. Given the delay embedding representation, we tested the Sparse Regression (SR) method \cite{brunton_discovering_2016} on a second order polynomial representation of the delay embedding states. The time integration of the SR model is carried using the LOSDA solver \cite{odepack}. Regarding deep learning models, we compare our method to a stacked Bidirectional LSTM (RNN) and to the latent-ODE model \cite{chen2018neural}. The RNN model includes 4 LSTM layers with 10 hidden units. The input sequence of the model is an observation sequence of size 30. The Latent-ODE model involves a recognition network, a decoder and a dynamical model. The recognition network is an RNN with 40 hidden units. We use a 3-dimensional latent space that evolves in time according to a linear quadratic neural ODE model. The decoder network is a fully-connected neural network with one hidden layer and 15 hidden units.

The proposed framework is tested with a dimension of the augmented state space $d_E = 3$, {\em i.e.}, two latent states are concatenated to the observed variable. Overall the augmented state $\vect{u}$ is constructed as $\vect{u}_t = [\vect{z}_{1,t},\vect{y}_{1,t},\vect{y}_{2,t}]^T$. The parameters of the data-driven model governing $\vect{u}$, as well as the latent states $\vect{y}_{1,t}$ and $\vect{y}_{2,t}$ are optimized based on both i) the initial NbedDyn formulation presented as proposed in \cite{ouala2020learning} ({\em i.e.}, using the optimization criterion given by \eqref{eq:C5_opti_crit 2}) and ii) the constrained version introduced in this work, illustrated by criterion \eqref{eq:C5_opti_crit_const}. 

\begin{table*}[h]
\begin{adjustbox}{width={\textwidth},totalheight={\textheight},keepaspectratio}
\centering
\begin{tabular}{lll*{4}c}
\toprule
\multicolumn{2}{c}{Model} &$t_0+dt$ &$t_0+4dt$  & $\lambda_1$ & $\lambda_1$ \\
\midrule \midrule 
\multirow{4}{*}{SR}
%\addlinespace \\
& $\tau_{MI} =16$, $d_E(FNN) = 3$ &$2.8E-2$  & $1.4E-1$ &$NaN$ &$NaN$  \\
& $\tau_{MI} =16$, $d_E$(Whitney)$ = 6$ &$1.3E-1$  & $3.0E-1$ &$NaN$ &$NaN$  \\
& $\tau_{Corr} =27$, $d_E$(FNN)$ = 3$ &$6.6E-1$  & $2.9$ &$NaN$ &$NaN$  \\
& $\tau_{Corr} =27$, $d_E$(Whitney)$= 6$ &$1.9E-1$  & $9.0E-1$ &$NaN$ &$NaN$  \\
& $\tau = 6$, $d_E = 3$ &$4.0E-3$  & $0.02.7E-2$ &$NaN$ &$NaN$  \\
& $\tau = 10$, $d_E = 3$ &$0.01.3E-2$  & $0.07.3E-2$ &$NaN$ &$NaN$  \\
\midrule
\multirow{1}{*}{Latent-ODE}
&    &$6.5E-2 \pm 4.4E-2$  & $1.5E-1 \pm 1.1E-1$ &$NaN$ &$NaN$  \\
\midrule
\multirow{1}{*}{RNN}
&    &$2.7E-1\pm5.7E-2$  & $3.1E-1\pm1.4E-1$ &$-5.4\pm0.0$  \\
\midrule
\multirow{1}{*}{NbedDyn}
&   $d_E = 3$ & \textbf{8.8E-5} $\pm$ 2.9E-5 & 5.8E-4 $\pm$ 1.2E-4 & 0.88 $\pm$ 0.02  & $NaN$ \\
\multirow{1}{*}{Constrained NbedDyn}
&   $d_E = 3$ & 1.2E-4 $\pm$ 4.1 E-5 & \textbf{3.5E-4} $\pm$ 7.4 E-5 &  \textbf{0.91} $\pm$ 0.02 & \textbf{0.90} $\pm$ 0.02 \\

\bottomrule
\end{tabular}
\end{adjustbox}
\caption {{\bf  \em Forecasting performance on the test set of the benchmarked data-driven models for Lorenz-63 case-study}: first two columns : root mean square error (RMSE) for different forecasting time-steps, third column : largest Lyapunov exponent of a predicted series of length of 10000 time-steps (The true largest Lyapunov exponent of the Lorenz 63 model is 0.91 \cite{Sprott_chaos}).}
\label{tab:C5_for_63}
\end{table*}

\begin{figure*}
\centering
\includegraphics[width=1.0\textwidth,height=11.0cm]{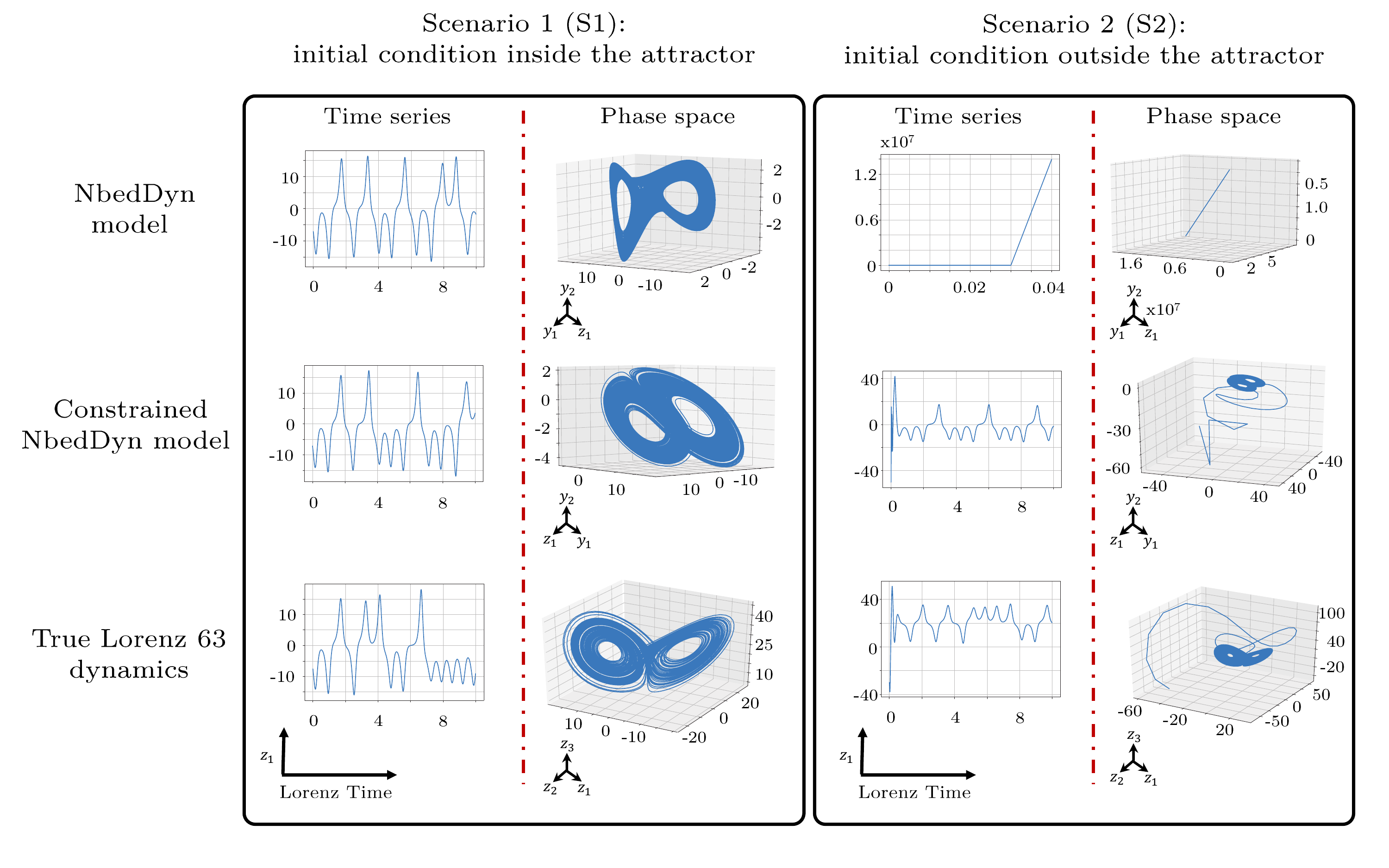}
\caption{{{\bf   \em Forecasting performance of the NbedDyn models for Lorenz 63 case-study.} First row, NbedDyn model as presented in \cite{ouala2020learning}; second row, proposed constrained NbedDyn model; third row, true Lorenz 63 model. The left (resp. right) panel illustrates the forecasting performance of the NbedDyn for an initial condition inside (resp. outside) the attractor. It is worth noting that the scales of the forecasted latent states $\vect{y}_1$ and $\vect{y}_2$ are different from the ones of the unseen states $\vect{z}_1$ and $\vect{z}_2$. This stems from the fact that given partial observations of the system, several models can reflect the variability of the observations while being diffeomorphic to the actual governing dynamics.}}
\label{fig:C5_forecast_lor63}
\end{figure*}

\textbf{{Forecasting performance of the proposed data-driven models:}} We evaluate in Table~\ref{tab:C5_for_63} the short-term forecast as well as the topological structure of the simulated limit-sets (illustrated for instance through the largest Lyapunov exponent). Regarding the short-term forecast, both versions of the NbedDyn model outperform benchmarked state-of-the-art techniques in terms of root mean square error (RMSE). The constrained version results in a slightly larger RMSE score due to the additional boundedness terms in the optimization criterion. In the other hand, when considering the long term simulation performance of the tested data-driven models, when the initial condition is inside the spanned attractor of the augmented states, the dynamical model optimized using the criterion (\ref{eq:C5_opti_crit 2}) gives trajectories that are bounded and with similar topological characteristics to the true Lorenz 63 model. However, when the initial condition is far from the spanned attractor, the model optimized through (\ref{eq:C5_opti_crit 2}) diverges to infinity. From a machine learning perspective, this is the direct consequence of a poor generalization performance to states that are far from the attractor spanned by the training data. In this situation, our model involves several attracting regions of chaotic and unstable solutions. When the initial condition is far from the spanned attractor, the state evolution is dominated by a positive energy growth, which makes our model diverge to infinity. The constrained model in the other hand, satisfies elementary conservation constraints that are present in the actual Lorenz 63 system and leads to a bounded behavior with a large attracting region of the chaotic limit-set. We further illustrate these conclusions with forecasting examples depicted in Figure~\ref{fig:C5_forecast_lor63} with initial conditions lying both inside and outside the attractor. 

\subsection{Lorenz 96}
The Lorenz-96 system \cite{lorenz1996predictability} with periodic boundary conditions is governed by the following ODE:
\begin{equation}
\label{eq:lorenz-96}
\frac{d\vect{z}_{t,i}}{dt}=(\vect{z}_{t,i+1}-\vect{z}_{t,i-2})\vect{z}_{t,i-1} - \vect{z}_{t,i} +F
\end{equation}
with $\vect{z}_{t,-1}=\vect{z}_{t,s}$, $\vect{z}_{t,s+1}=\vect{z}_{t,1}$.

The Lorenz-96 state sequences is also simulated using the LOSDA ODE solver \cite{odepack} with $F = 8$ and a dimension $s = 40$. Similar to the previous experiment, the numerical simulation consists of $5000$ time steps sampled at $dt = 0.01$. The initial $4000$ measurements of the first 20 states, {\em i.e.}, $\vect{x}_t = \mathcal{H}(z_t) = [\vect{z}_{1,t}, \cdots, \vect{z}_{20,t}$]$^T$ are used as training data. The remaining time steps are used as a test set. In this experiment, the projection operator in \eqref{eq:C5_aug_vect} is the identity matrix, {\em i.e.} $\mathcal{M(\cdot)} = I_{20}$.

\textbf{Parametrization of the data-driven models} : The proposed framework, is tested with a dimension of the augmented state space $d_E = 40$ {\em i.e.}, 20 latent states are concatenated to the observed variable. In this experiment, the quadratic part of the NODE model $f_{\theta}$ is convolutional and corresponds to the true non-linear interactions in \eqref{eq:lorenz-96} with trainable weights. Such representation significantly accelerates the computation of the constraint $\mathcal{C}_1$ in \eqref{eq:C5_opti_crit_const} \footnote{Other experiments, not shown here, demonstrate that a full linear quadratic NODE model also converges (but at a slower rate) to solutions with the same properties as the convolutional model evaluated in this experiment.}. Similar to the Lorenz 63 experiment, both the constrained and unconstrained versions of the model are compared to the sparse regression technique, to a stacked bidirectional LSTM (RNN) and to the Latent-ODE model \cite{chen2018neural}). The SR model is based on a second-order polynomial representation. The state-space of the model is built on a stacked delay embedding of the 20 observed states. For each observed state, the dimension of the embedding is computed using the FNN method and the lag is chosen according to both the mutual information and the correlation techniques. The time integration of the SR model is carried using the LOSDA solver \cite{odepack}. The RNN model includes 10 LSTM layers with 100 hidden units. The input sequence of the model is an observation sequence of size 20. Regarding the Latent-ODE model, the recognition network is an RNN with 100 hidden units. The dimension of the latent space is set to 40 and the dynamical model is a linear quadratic neural ODE. The decoder network is a fully-connected neural network with one hidden layer and 100 hidden units.

\begin{table*}[htb!]
\begin{adjustbox}{width={\textwidth},totalheight={\textheight},keepaspectratio}
\centering
\begin{tabular}{ll*{4}c}
\toprule
\multicolumn{2}{c}{Model} &$t_0+dt$ &$t_0+4dt$  & $\lambda_1$ & $\lambda_1$ \\
\midrule \midrule 
\multirow{1}{*}{SR( $\tau_{MI}$ )}
&     & $>10$  & $>10$ &$NaN$ &$NaN$  \\
\multirow{1}{*}{SR( $\tau_{Corr}$ )}
&  & $>10$  & $>10$ &$NaN$ &$NaN$  \\
\midrule
\multirow{1}{*}{Latent-ODE}
&   &  0.262 $\pm$ 2.08E-1 &  0.560 $\pm$ 3.7E-1 & NaN & NaN   \\
\midrule
\multirow{1}{*}{RNN}
&   &   0.186 $\pm$ 3.6E-2 & 0.231 $\pm$ 5.7E-2 & 5.091 $\pm$ 1.856 & 2.451 $\pm$ 2.092  \\
\midrule
\multirow{1}{*}{NbedDyn}
 &   & \textbf{0.009} $\pm$ 2.4E-3  & 0.293 $\pm$ 6.4E-2 & NaN & NaN  \\
\multirow{1}{*}{Constrained NbedDyn}
 &   & 0.012 $\pm$ 2.3E-3  & \textbf{0.036} $\pm$ 6.9E-3 & \textbf{1.211} $\pm$ 1.32E-1 &\textbf{ 1.399 }$\pm$ 6.6E-2  \\

\bottomrule
\end{tabular}
\end{adjustbox}
\caption {{\bf  \em Forecasting performance on the test set of the data-driven models for Lorenz-96 dynamics where only the first 20 state variables are observed}: first two columns : mean RMSE for different forecasting time-steps, third column : largest Lyapunov exponent of a predicted series of length of 10000 time-steps (The true largest Lyapunov exponent of the Lorenz 96 model is 1.67 \cite{brajard2020combining}).}
\label{tab:C5_for_96}
\vspace{-0.3cm}
\end{table*}

\textbf{{Forecasting performance of the proposed data-driven models:}}
Similar to the Lorenz 63 experiment, both versions of the NbedDyn model outperform benchmarked state-of-the-art techniques in terms of short-term forecast RMSE. When considering the attractor reconstruction based on long-term simulation of the data-driven models, the constrained version of the NbedDyn model is the only model able to unfold the Lorenz 96 attractor. The other data-driven models, including the unconstrained NbedDyn either diverge to infinity or generate trajectories that do not match the Lorenz 96 hidden attractor (illustrated through the largest Lyapunov exponent).
\begin{figure}
\centering
  {
\includegraphics[width=0.5\textwidth]{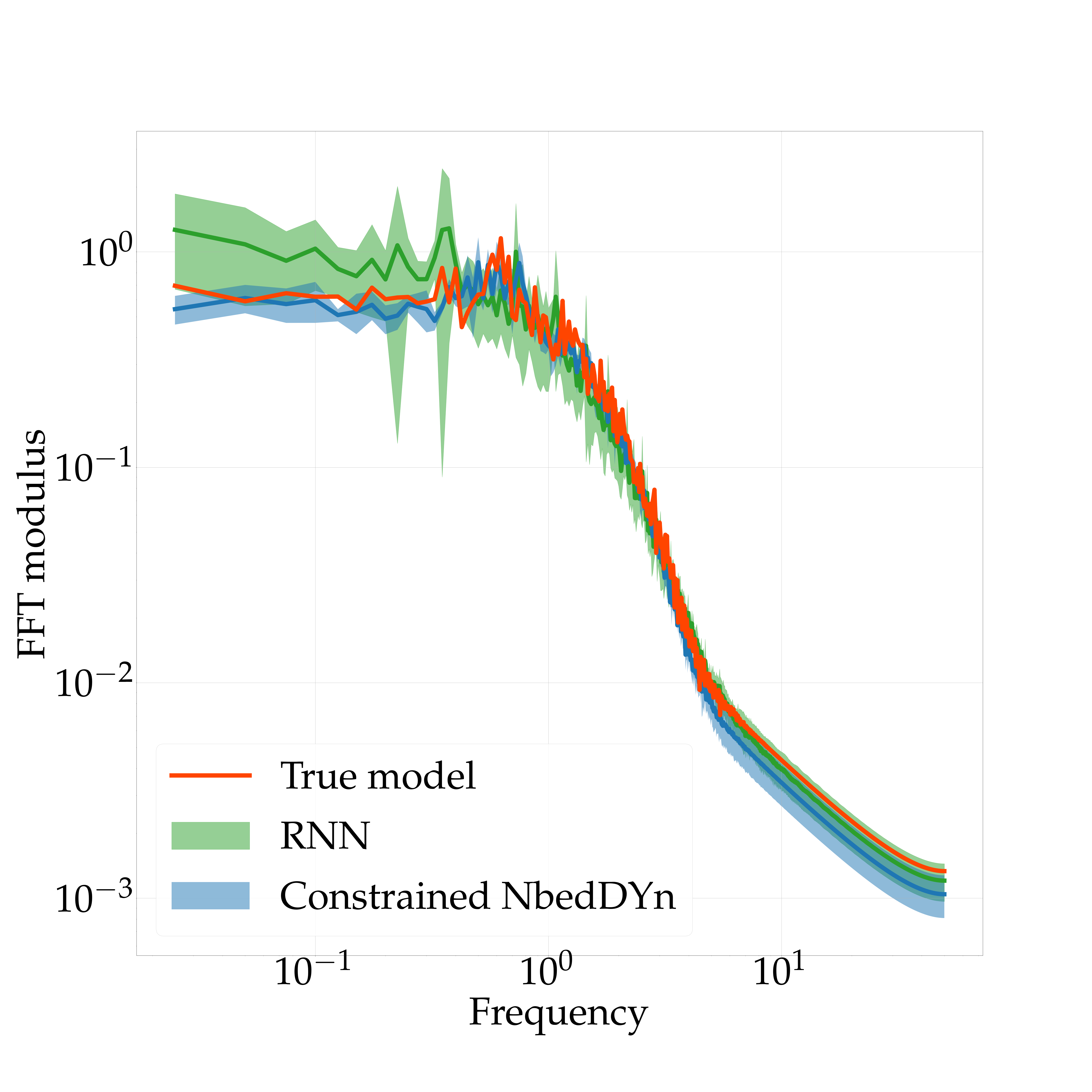}}
\caption{{{{ \bf  \em Mean Fast Fourier Transform (FFT) modulus of the benchmarked data-driven models for Lorenz-96 case-study}. We report the FFT modulus of simulated time series over the test period for the benchmarked data-driven models. The solid lines represent the FFT averaged over five training runs for each scheme. The light-color interval is the standard deviation of the different models. The unconstrained NbedDyn trajectories diverge after a short forecasting time, thus, its FFT is omitted.}
}}
\label{fig:C5_PSD_L96}
\end{figure}

\begin{figure*}
\centering
\includegraphics[width=\textwidth]{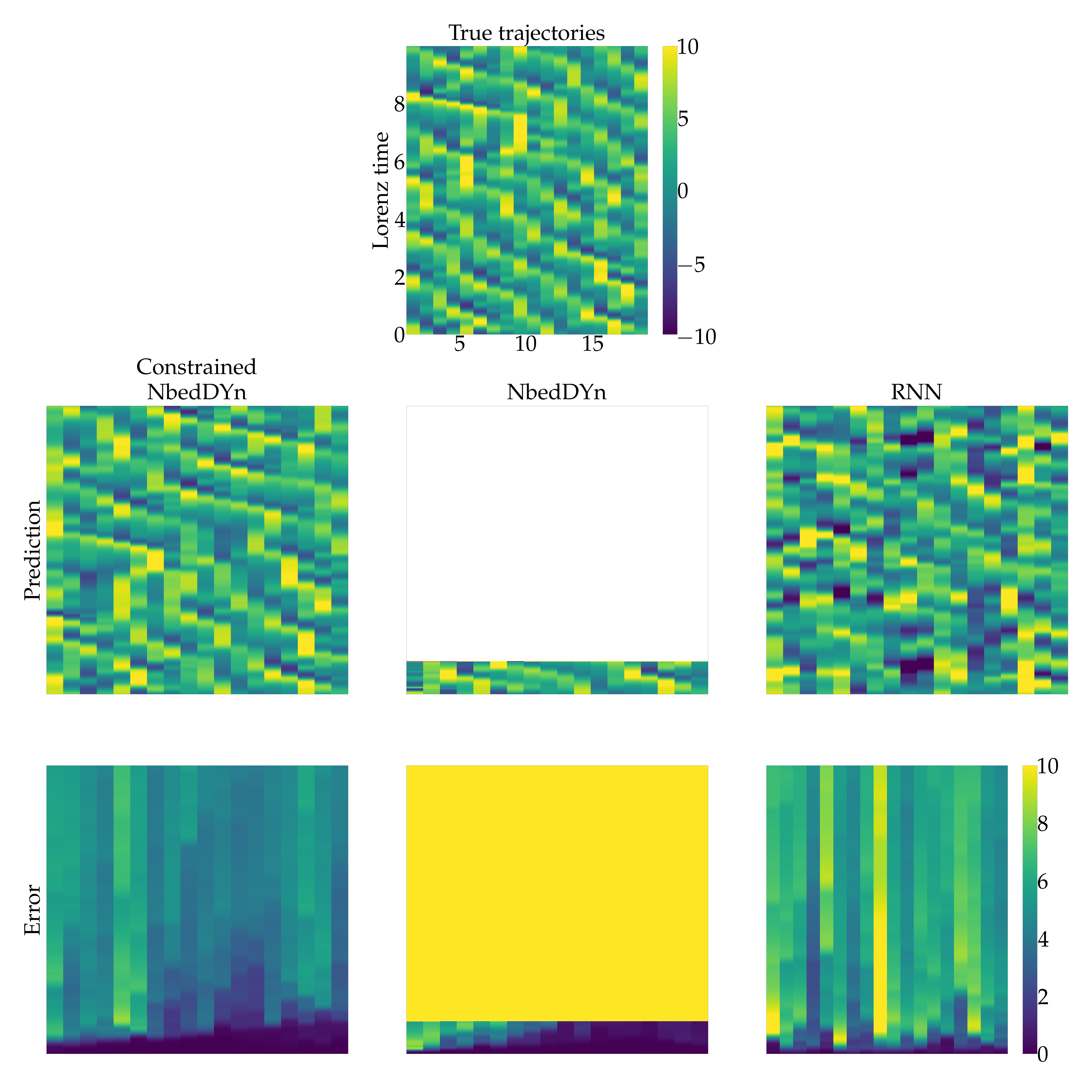}

\caption{{{{ \bf  \em Forecasting performance on the benchmarked data-driven models for Lorenz-96 case-study}. First row: ground-truth Lorenz 96 time series; Second row : data-driven model simulations over a test period; Third row: associated RMSE map. For each subplot, the vertical axis refers to the time axis and the horizontal axis to the 20-dimensional observation state. As unconstrained NbedDyn trajectories blowup after a short time horizon, no observation state can be depicted. }
}}
\label{fig:C5_forecast_L96}
\end{figure*}

We further illustrate the quantitative analysis conclusions through the visual comparison of the FFT modulus as well as a forecasting example in figures \ref{fig:C5_PSD_L96} and \ref{fig:C5_forecast_L96} respectively. The proposed architecture shows a better match to the true trajectory both in the temporal and spectral domains. Specifically, the constrained NbedDyn trajectory, although diverges from the ground truth (due to the chaotic nature of the attractor), keeps a similar spatio-temporal behavior illustrated by the prediction example in Fig. \ref{fig:C5_forecast_L96} as well as the spectrum in Fig. \ref{fig:C5_PSD_L96}. Interestingly, this performance is truly due to the boundedness constraints since the unconstrained version, and even using the same ODE parameterization, diverges to infinity after a short simulation time. The RNN trajectories decently capture the time frequency patterns of the true states but fail to reproduce the Lorenz 96 dynamics in the phase space.

\subsection{Patch shallow Water equation (PSWE)}
As a third case-study, we consider SWE dynamics governed by the following set of equations:
\begin{equation}
\left\{
\begin{aligned}
&\frac{\partial v_x}{\partial t} - F_{v_x} = -g \frac{\partial \eta}{\partial x}\\
&\frac{\partial v_y}{\partial t} - F_{v_y} = -g \frac{\partial \eta}{\partial y}\\
&\frac{\partial \eta}{\partial t} + \frac{\partial (\eta +H)v_x}{\partial x} +\frac{\partial (\eta +H)v_y}{\partial y}  &= 0
\label{eq:SWE_dynamics}
\end{aligned}
\right.
\end{equation}
where $x$ and $y$ represent the 2D directions of the fluid. $\eta$ is the fluid surface elevation, and $(v_x,v_y)$ is the fluid's horizontal flow velocity. g is the acceleration due to gravity, which is taken to be equal to $9.81 m/s^2$. $H$ corresponds to the total depth (here $H = 100 m$) of the fluid and $f = f_0 + \lambda y$ is the Coriolis forcing where $f_0 = 1E-4 s^{-1}$ and $\lambda = 2E^{-11}(m\times s)^{-1}$.

The direct numerical simulation of \eqref{eq:SWE_dynamics} is carried using a finite difference method. The length of the domain is set to $1000 km \times 1000 km$ with a corresponding regular discretization of $80 \times 80$. The temporal step size $dt$ is set to satisfy the Courant–Friedrichs–Lewy condition ($dt=40.41$ seconds). As observed process, we consider a low-resolution version of a patch of size $250km \times 250km$ of the sea surface elevation $\eta$. The patch is located in the center of the 2D domain. The low-resolution projection is computed from an 8-dimensional EOF decomposition, which amounts to capture $80\%$ of the total variance. Regarding the training configuration, $100000$ simulation time steps are generated. The transient (first $2500$) time steps are omitted, and we use the post-transient first 49701 time-steps as training data. The remaining sequence is used as a test set. %we use the first 49701 time-steps. The remaining time-steps define the test data.
%\todo[inline]{Lire Please, la partie importance du cas PSWE}

This case study combines numerous layers of complexity when compared to the previous experiments. Specifically, the dynamics of the considered $250km \times 250km$ domain depend on the neighboring regions. Its variability  is rigorously governed by the Shallow Water Equation in the full domain. As such, one cannot derive an autonomous differential system for the subdomain. Besides, through the EOF-based filtering, the observed process only refers to coarse-scale dynamics. Reduced-order modeling schemes \cite{quarteroni2015reduced} and learning-based closure models \cite{frezat2021physical} would be typical solutions to address this issue, which would rely on the approximation of the observed dynamics by an autonomous system in the observation space or in a lower-dimensional projection of this space. 
We show in this experiment that the proposed framework can account for unknown boundary conditions and fine-scale dynamics by retrieving the most appropriate missing information to forecast future coarse-scale observations within the considered patch. We may remind that the learning of the latent states only relies on the observed variables.

\textbf{Parametrization of the data-driven models} : All the tested models are built on the 8-dimensional EOF basis used to filter the observations. In the proposed framework, this setting corresponds to a specific parameterisation of the projection operator $\mathcal{M(\cdot )}$ in \eqref{eq:C5_aug_vect} where $\mathcal{M}(\vect{x}_t) \in \mathbb{R}^8$ is an EOF projection. The NbedDyn model (both the constrained and unconstrained settings) is constructed with a dimension of the augmented state $d_E = 18$. Our framework is compared to a stacked Bidirectional LSTM (RNN) and to a linear regression model. The RNN includes 10 LSTM layers with 100 hidden units. The input sequence of the RNN is an observation sequence of size 40. The linear model, that we call here Delay EDMD, is built on a Singular Value Decomposition (SVD) of delay embedding coordinates of the observations \cite{kamb2020timeDobsKoopman}. The delay embedding is computed, for every EOF component, using a lag embedding equal to one time step and a dimension equal to 200. The dimension of the SVD is set to 150, which accounts for over 99.99\% of the total variance of the delay embedding representation.

\begin{table*}[h]
\centering
\begin{tabular}{lll*{4}c}
\toprule
\multicolumn{3}{c}{Model} &$t_0+dt$ &$t_0+4dt$  &  &  \\
\midrule \midrule 
\multirow{1}{*}{RNN}
&   &  & 7.35E-4 $\pm$ 2.55E-4 & \textbf{1.28E-3} $\pm$ 3.71E-4 &  &   \\
\midrule
\multirow{1}{*}{NbedDyn}
 &   &   & 1.33E-3 $\pm$ 7.88E-5 & 3.66E-3 $\pm$ 1.92E-4 & &  \\
 \midrule
\multirow{1}{*}{Delay EDMD}
 &   &   & 1.43E-3  & 1.42E-3 & &  \\
\midrule
\multirow{1}{*}{Constrained NbedDyn}
 &   &   &\textbf{6.28E-4 }$\pm$ 1.69E-4  & 1.76E-3 $\pm$ 4.7E-04 & & \\

\bottomrule
\end{tabular}
\caption {{\bf  \em Forecasting performance of the benchmarked data-driven models for PSWE case-study}: we report the mean RMSE mertrics for different forecasting time-steps for the tesst set.}
\label{tab:C5_for_PSWE}
\vspace{-0.3cm}
\end{table*}

\begin{figure*}
\centering
\includegraphics[width=\textwidth]{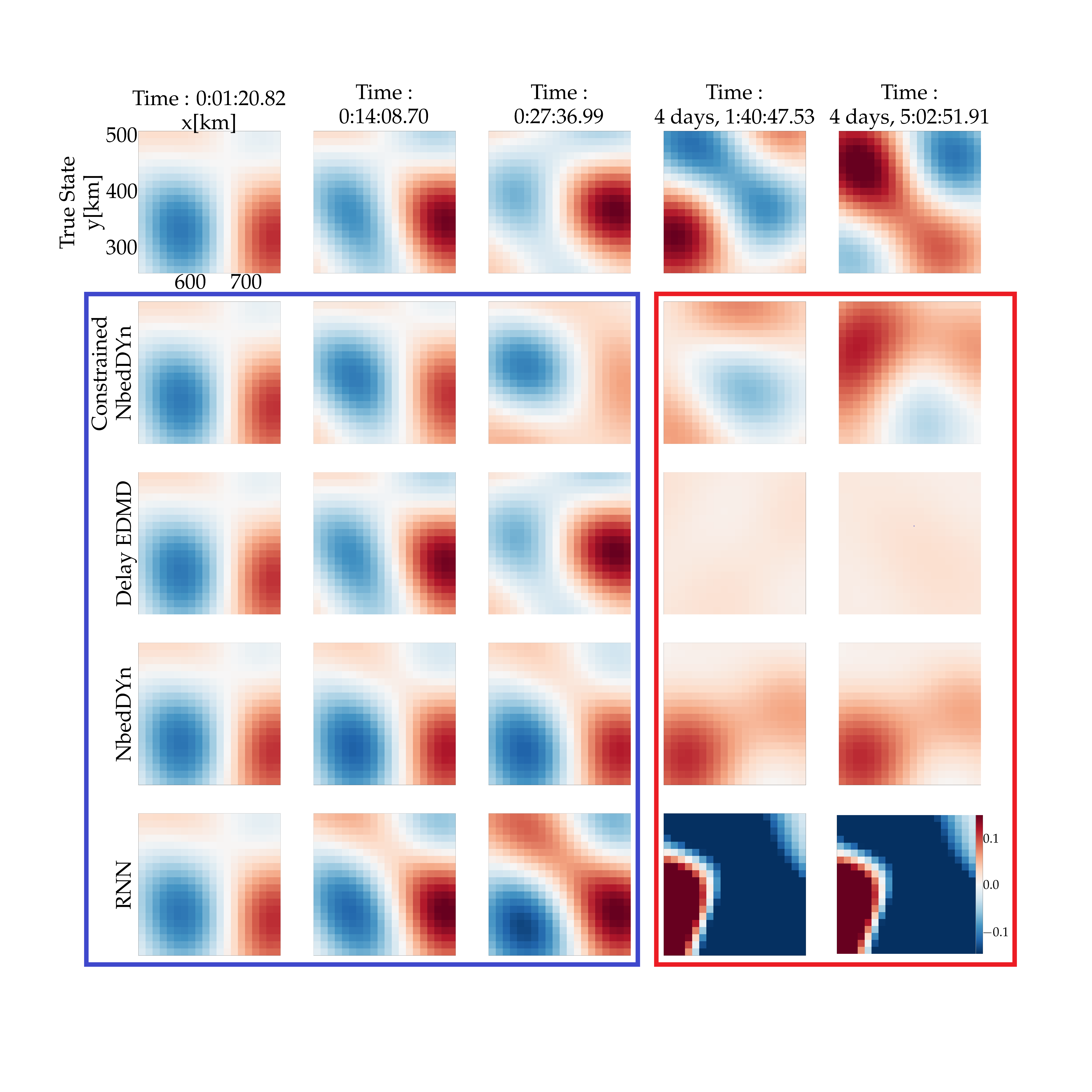}
\caption{{{{\bf \em Prediction example over the test period of the data-driven models for the PSWE case-study}. The blue panel illustrates a short-term forecast for the benchmarked models and the red panel highlights the associated long-term simulations. A spectral analysis of the simulated trajectories, both in space and time, are given in figure \ref{fig:C5_PSD_SWE}.}
}}
\label{fig:C5_forecast_patchs_SWE}
\end{figure*}

\begin{figure*}
\centering
  \subfloat[]{%
\includegraphics[width=0.5\textwidth]{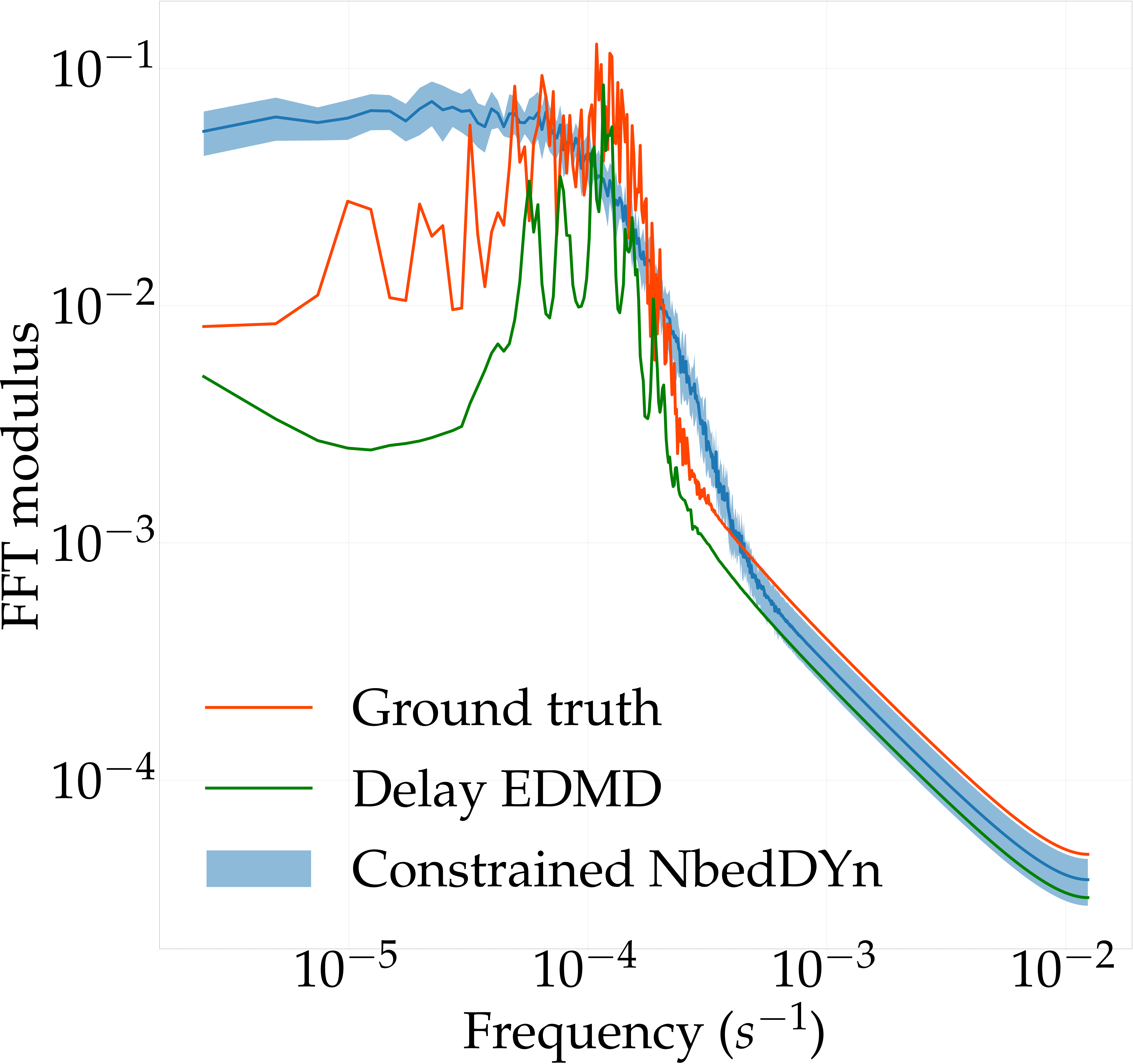}}
  \subfloat[]{%
\includegraphics[width=0.5\textwidth]{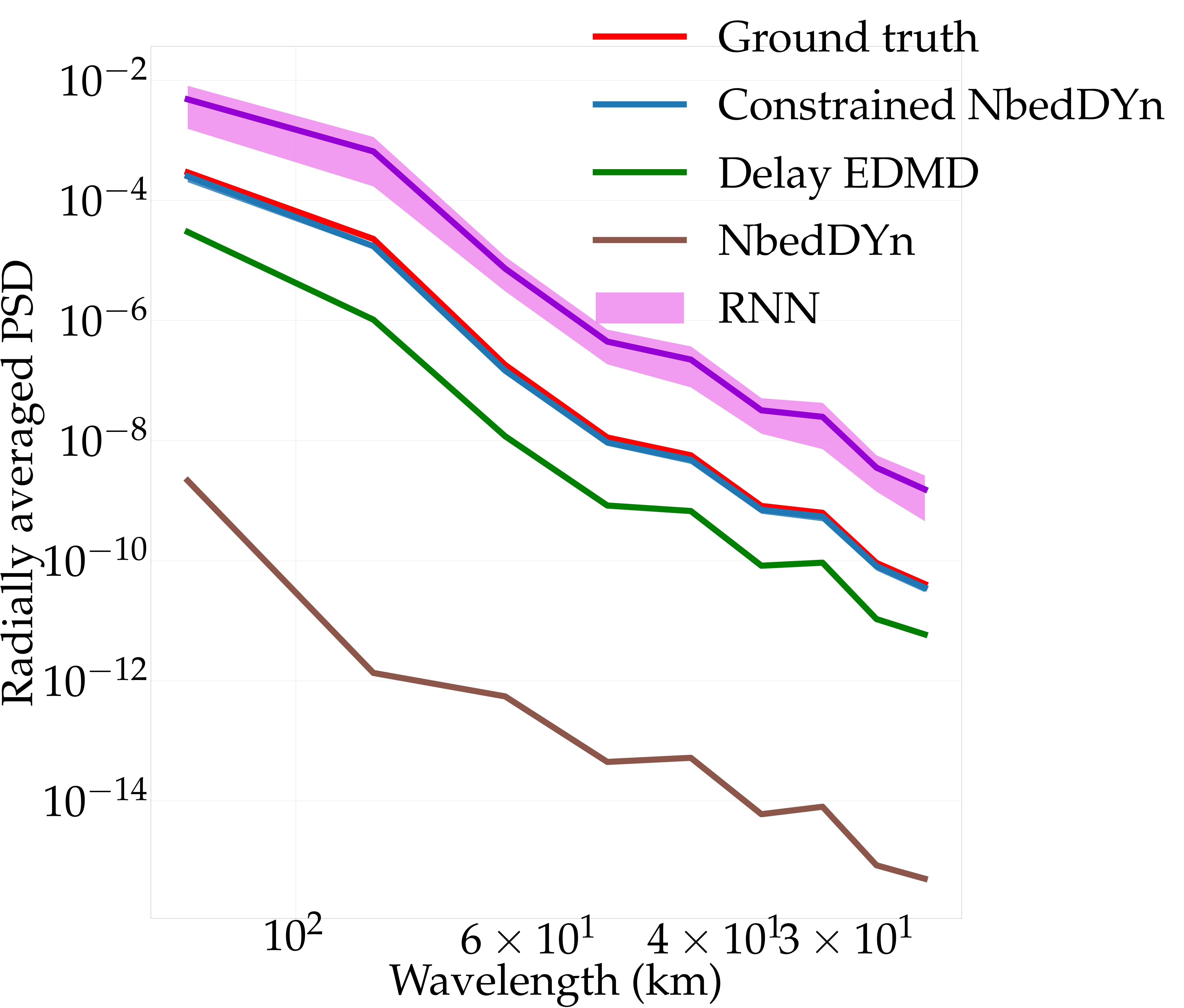}}
\caption{{{{ \bf  \em Spatio-temporal spectrums of the data-driven models with respect to the ground truth}. The FFT modulus of the data-driven models in (a) are computed from a simulated time series in the test set. The radially-averaged Power Spectral Density (PSD) in (b) is averaged over a simulation of size 10000 time steps .The lines represent the averaged spectrums, computed over five optimized models. The light-color intervals refer to the standard deviation.}
}}
\label{fig:C5_PSD_SWE}
\end{figure*}

\textbf{{Forecasting performance of the proposed data-driven models:}} regarding the short-term forecasting performance, reported for instance in table~\ref{tab:C5_for_PSWE}, the benchmarked models achieve similar errors. These results are also highlighted in the forecasting examples given in figure \ref{fig:C5_forecast_patchs_SWE} where the short term forecast, highlighted in the blue panel, are similar to the true state, up to 27 minutes where only the Delay EDMD technique is still qualitatively close to the ground truth. The analysis of the long-term simulated states, bordered by a red panel in Figure \ref{fig:C5_forecast_patchs_SWE}, gives in the other hand a different conclusion. While the proposed constrained NbedDyn model keeps simulating realistic states, both the unconstrained NbedDyn and RNN models get stuck either at equilibrium points, or at very slow manifolds. The linear Delay EDMD model converges to the origin after a finite amount of time, which is expected since finite-dimensional linear models can not reflect a chaotic behavior. We can draw the same analysis from the mean FFT modulus of the forecasted EOF modes, as well as the mean radially averaged PSD of the 2D fields in Fig.~\ref{fig:C5_PSD_SWE}. When considering the spatial spectrum, only the proposed framework is able to match perfectly the spectrum of the ground-truth. This result illustrates the ability  of our model to capture all the spatial scales of the dynamics and to keep simulating these scales after long simulation times. Regarding the temporal spectrum, the constrained NbedDyn and the Delay EDMD models match the spectrum of the ground truth both in the medium and high frequency range. The low-frequency range in the other hand is not well captured by the model. This suggests that we may need to consider an optimization objective \eqref{eq:C5_opti_crit_const} with longer prediction times than the one used in this work to embed realistic low-frequency patterns.

\section{Conclusion}
\label{sec:Conclusion}
Two central challenges in data-driven modeling involve how to handle partial observations and how to guarantee stability in the resulting models.  
In this paper, we demonstrate that combining an augmented neural ODE formulation and boundedness constraints, it is possible to considerably improve the data-driven identification and forecasting of partially-observed dynamics. 
{The proposed framework is assessed on several case studies with respect to state-of-the-art forecasting and system identification techniques}. Specifically, for classical chaotic ODEs (Lorenz 63 and 96), the proposed methodology matches state-of-the-art short-term forecasting performances while ensuring realistic long-term patterns for partially-observed settings. We also report a more complex case study, in which we have considered low-resolution observations of a subdomain of SWE dynamics with unknown boundary conditions. The results from this study support the potential of the proposed technique to derive faithful data-based representations of ocean flows, especially ocean surface dynamics. 
Such dynamics involve, similar to the PSWE case study, but at a higher level of complexity, unseen components, missing forcing, unknown boundary conditions and unresolved spatio-temporal scales. In all numerical experiments, both the data-driven model formulation in an augmented space and the boundedness constraints are key features to capture the dynamics underlying the observations.

An avenue of promising future work will involve extending the proposed framework to other non-linear models through the generalization of the boundedness constraints. Foremost, polynomial models could be considered by imposing, similar to the quadratic terms of the proposed architecture, an energy-preserving non-linearity. Other non-linear architectures may still need further investigation, since the fluctuation energy may not be the appropriate Lyapunov function to consider. The recent works of \cite{manek2020learning} may be useful for finding Lyapunov functions that guarantee stability and boundedness. 

To recall, explicitly constraining chaos within a data-driven formulation is rigorously intractable since such a behavior is only revealed, and thus potentially constrained using long-term simulations. From this point of view, the proposed framework is not guaranteed to reach an expected chaotic evolution, since this is not explicitly constrained within the framework. Specifically, when considering the bounded NbedDyn model, the learning criterion allows for any bounded limit-set, including stable ones, as long as the short-term forecast of the observations is minimized. This criterion does not guarantee a replication of the chaotic observations, and long-term simulations of the model can lead to undesirable stable limit-sets. Interestingly, stable limit-sets can be fully characterized by the dynamical equation, without resorting to brute force long-term simulation of the model. 
Thus, these limit sets can be propagated to infinity. However, sending stable limit-sets to infinity may result in new questions about how observation data may help to regularize the learning stage.

\section*{Acknowledgements} \label{sec:acknowledgements}
This work was supported by LEFE program (LEFE MANU project IA-OAC), CNES (grant OSTST DUACS-HR) and ANR Projects Melody and OceaniX. It benefited from HPC and GPU resources from Azure (Microsoft Azure grant) and from GENCI-IDRIS (Grant 2020-101030). SLB acknowledges valuable discussions with Alan Kaptanoglu. 

\bibliographystyle{unsrt}
\bibliography{references}

\end{document}